\documentclass{article}

\usepackage{iwpt97}

\makeatletter
\def\fnum@figure{\bf\figurename~\thefigure}
\def\fnum@table{\bf\tablename~\thetable}
\makeatother

\def\figref#1{Figure~\ref{#1}}
\def\tabref#1{Table~\ref{#1}}

\makeatletter
\def\shift{\mathop{\operator@font shift}}
\def\attach{\mathop{\operator@font attach}}
\def\project{\mathop{\operator@font project}}
\makeatother

\begin{document}
\title{PROBABILISTIC PARSING\\ USING LEFT CORNER LANGUAGE MODELS}
\author{Christopher D. Manning \\ 
        \large Linguistics F12 \\[-0.3ex] \large University of Sydney
        NSW 2006 \\[-0.3ex] \large Australia \\[-0.3ex]
	{\normalsize\sf cmanning@mail.usyd.edu.au}
	\AND
        Bob Carpenter \\ \large Lucent Technologies Bell Labs \\[-0.3ex]
        \large 600 Mountain Avenue, Room 2D-329 \\[-0.3ex]
	\large Murray Hill NJ 07974 \\[-0.3ex]
        {\normalsize\sf carp@research.bell-labs.com}}
\maketitle

\begin{abstract}
We introduce a novel parser based on a probabilistic version of a
left-corner parser.  The left-corner strategy is attractive because
rule probabilities can be conditioned on both top-down goals and
bottom-up derivations.  We develop the underlying theory and explain
how a grammar can be induced from analyzed data.  We show that the
left-corner approach provides an advantage over simple top-down
probabilistic context-free grammars in parsing the {\em Wall Street
Journal} using a grammar induced from the Penn Treebank.  We also
conclude that the Penn Treebank provides a fairly weak testbed due to
the flatness of its bracketings and to the obvious overgeneration and
undergeneration of its induced grammar.
\end{abstract}

\section{Introduction}

For context-free grammars (CFGs), there is a well-known
standard\footnotetext{We thank Edward Stabler and Mark
  Johnson for getting us interested in left corner parsing, and Mark more
  particularly for valuable discussion of some of the work in this
  paper.} 
probabilistic version, Probabilistic Context-Free Grammars (PCFGs), which
have been thoroughly investigated 
\cite{Suppes70,Sankoff71,Baker79,LaYo90,Kupiec91,JeLaMe92,Charniak93}.

Under this model, one assigns probabilities for
different rewrites of a non-terminal.  Or in other words, one is giving the
probability of a local subtree given the mother node.  So, for example, we
might have:
%
\[
P(NP \rightarrow Det\ N|NP \mbox{ mother})  = 0.2 \hspace*{1in} 
P(NP \rightarrow Pron|NP \mbox{ mother})  =  0.1
\]
where in general, $\forall$ nonterminals $A$,
$\sum_{\gamma} P(A \rightarrow \gamma) = 1$.

But standard PCFGs are only one way to make a probabilistic version of CFGs.
If we think in parsing terms, a PCFG corresponds to a probabilistic version
of top down parsing, since at each stage we are trying to predict the child
nodes 
given knowledge only of the parent node. 
Other parsing methods lend
themselves to different models of probabilistic conditioning.
Usually, such conditioning is a mixture of top-down and bottom-up
information.  This
paper discusses some initial results from another point in this
parameter space where the conditioning reflects 
a left-corner parsing strategy, yielding what we will call
probabilistic left-corner grammars (PLCGs).\footnote
  {This name may appear strange since the symbolic part of the grammar
   is unchanged and still context-free.  But if we regard the
   probabilistic conditioning as part of the grammar, then we do have a
   different kind of grammar.  We should then perhaps call the result a
   LCPG, but we place the P in initial position for reasons of tradition.}
Left-corner parsers simultaneously
work top-down from a goal category and bottom-up from the left corner
of a particular rule.  For instance, a rule such as $S \rightarrow NP\
VP$ has the left-corner $NP$ and will be fired whenever an $NP$ has
been derived and an $S$ would help toward the eventual goal category.
In this paper, we present algorithms
for PLCG parsing, present some results comparing PLCG parsing with
PCFG parsing, and discuss some mechanisms for improving results.

Why might one want to employ PLCGs?  While the main perceived weakness of
PCFGs is their lack of lexicalization, they are also
deficient on purely structural grounds \cite{BrCa93}.
Inherent to the idea of a PCFG is that probabilities are context-free: for
instance, that the probability of a noun phrase expanding in a certain way
is independent of where the NP is in the tree.  Even if we in some way
lexicalize PCFGs to remove the other deficiency, this assumption
of structural context-freeness remains.
But this context-free assumption is actually quite wrong.  For
example, \tabref{notcf} shows how the probabilities of expanding an NP node
(in the Penn Treebank) differ wildly between subject position and object
position.  Pronouns, proper names and definite NPs appear more commonly in
subject position while NPs containing post-head modifiers and bare nouns
occur more commonly in object position (this reflects the fact that the
subject normally expresses the sentence-internal topic \cite{Manning96}).
\begin{table}
\begin{center}
{\small
\begin{tabular}{lrr}
Expansion & \% as Subj & \% as Obj\\[2ex]
NP $\rightarrow$ PRP & 13.7\% & 2.1\% \\
NP $\rightarrow$ NNP & 3.5\% & 0.9\% \\
NP $\rightarrow$ DT NN & 5.6\% & 4.6\% \\
NP $\rightarrow$ NN & 1.4\% & 2.8\% \\
NP $\rightarrow$ NP SBAR & 0.5\% & 2.6\% \\
NP $\rightarrow$ NP PP & 5.6\% & 14.1\%
\end{tabular}}
\end{center}
\caption{Selected common expansions of NP as Subject vs.\ Object}\label{notcf}
\end{table}

Another advantage of PLCGs is that parse probabilities are
straightforwardly calculated from left to right, which is convenient for
online processing and integration with other linear probabilistic
models.%
\footnote{Note however, that while the obvious way of calculating PCFG
probabilities does not allow incremental processing, incremental
calculation is possible, as discussed by \cite{JeLaMe92}.}

\section{Probabilistic Left Corner Grammars}


Left corner parsers \cite{RoLe70,Demers77} work by a combination of
bottom-up and top-down processing.  One begins with a goal category
(the root of what is currently being constructed), and then looks at
the left corner of the string (i.e., one shifts the next terminal).
If the left corner is the same category as the goal category, then
one can stop.  Otherwise, one projects a possible local tree from the
left corner.  The remaining children of this projected local tree
then become goal categories and one recursively does left corner
parsing of each.  When this local tree is finished, one again
recursively does left-corner parsing with this subtree as the left
corner, and the same goal category.  To make this description more
precise, a Prolog version of a simple left corner recognizer is shown
in \figref{lcprog}.  This particular parser assumes a rule format for
{\tt rule}/2 that allows lexical material to appear on the right-hand
side of a rule.%
\footnote{In general, empty categories can be accommodated by allowing
a category to be introduced for completion without popping a word off
the input stack.}
\begin{figure}
{\small
\begin{verbatim}
% lc(List_of_words_to_parse)
lc(Ws) :- start(C),
    complete_list([C], Ws, []).

complete(C, C, Ws, Ws).           % attach
complete(W, C, Ws, NewWs) :-      % project lc
    rule(LHS, [W|Rest]),          % lex / phrase
    complete_list(Rest, Ws, Ws2),
    complete(LHS, C, Ws2, NewWs).

complete_list([], Ws, Ws).
complete_list([C|Cs], [W|Ws], NewWs) :- 
    complete(W, C, Ws, Ws2),   % shift
    complete_list(Cs, Ws2, NewWs).
\end{verbatim}}
\caption{A Prolog LC parser}\label{lcprog}
\end{figure}

A common formulation of left corner parsers is in terms of a stack of found
and sought constituents, the latter being represented as
minus categories on the stack (and represented as \verb+m(Cat)+ in the
Prolog code).  A left corner parser that uses a stack is shown in
\figref{lcstackprog}.
\begin{figure}
{\small
\begin{verbatim}
slc(Ws) :- start(C),
    slc(Ws, [m(C)]).

slc([], []).
slc(L0, Stack0) :-
    process(Stack0, Stack, L0, L),
    slc(L, Stack).

process([A, m(A)|Stack], Stack, L, L). % attach
process([Item|Items], Stack, L, L) :-  % project LC
    rule(LHS, [Item|Rest]),
    predict(Rest, [LHS|Items], Stack).
process(Stack, [L|Stack], [L|Ls], Ls). % shift

predict([], L, L).
predict([L|Ls], L2, [m(L)|NewLs]) :-
    predict(Ls, L2, NewLs).
\end{verbatim}}
\caption{A Prolog LC stack parser}\label{lcstackprog}
\end{figure}
Shifting is now an explicit option on a par with projecting and attaching,
but note that when to shift remains deterministic.
If the thing on top of the stack is a predicted
\verb+m(Cat)+, then one must shift, and one can never successfully shift at
other times.  This second version of the parser more transparently
corresponds to the probabilistic language model we employ.

To produce a language model that reflects the operation of a left corner
parser, we have to provide probabilities for the different operations
(the clauses of \verb+process+ in \figref{lcstackprog}).
For each step, we need to decide the probabilities of deciding to shift,
attach, or project.  The only
interesting choice here is deciding whether to attach in cases where the
left corner category and the goal category are the same.
For the other two operations of the parser, we need to model the
probability of shifting different terminals, and 
the probability of building a certain local tree given the left corner
($lc$) and the goal category ($gc$).  Under this model, we have 
probabilities for this last operation like this:
\begin{eqnarray*}
P(SBar \rightarrow P\ S|lc = P, gc = S) & = & 0.25\\
P(PP \rightarrow P\ NP|lc = P, gc = S) & = & 0.55
\end{eqnarray*}
How to make probabilities out of
the above choices is made precise in the next section.

\subsection{The LC probability of a parse}

In this section, we provide probabilities for left-corner
derivations.  These form the basis for a language model that assigns
probabilities to sentences.  
For a sentence $s$, we have that the probability
of a sentence according to a grammar $G$ is:
\begin{eqnarray*}
P(s|G) 
 & = & \sum_t P(s, t|G),\quad \mbox{$t$ a parse
tree of $s$}\\
& = & \sum_{\{t: {\rm\ yield}(t)=s\}} P(t|G)
\end{eqnarray*}
The last line follows since the parse tree determines the terminal yield.
It is therefore sufficient to be able to calculate the probability of a
(parse) tree.  Below we suppress the conditioning of the
probability according to the grammar.

Now following the intuition of our model having been inspired by left
corner parsing, we can express the probability of a parse tree in terms of
the probabilities of left corner derivations of that parse tree:
\[
P(t) = \sum_{d\ {\rm a\ LC\ derivation\ of}\ t} P(d)
\]
But under left corner parsing, each parse tree has a unique derivation and
so the summation sign can be dropped from this equation.

Now, without any assumptions, the probability of a derivation can be
expressed as a product in terms of the probabilities of each of the individual
operations in the derivation.  Suppose that $(C_1, \ldots, C_m)$ is the
sequence of operations in the LC parse derivation $d$ of $t$.  Then,
by the chain rule, we have:
\[
P(t) = P(d) = \prod_{C_1, \ldots, C_m}  P(C_i|C_1, \ldots, C_{i-1})
\]

In practice, we cannot condition the probability of each parse decision on
the entire history.  The simplest model, which we will explore for the rest
of this section, is to assume that the probability
of each parse decision is largely independent of the parse history, and
just depends on the state of the parser.  In particular, we will assume
that it depends simply on the left corner and top goal categories of the
parse stack.
This drastic assumption nevertheless gives us a slightly richer
probabilistic model than 
a PCFG, because elementary left-corner parsing actions are
conditioned by the goal category, rather than simply being the probability
of a local tree.  For instance, the probability of a
certain expansion of NP may be different in subject position and object
position, because the goal category is different.

Each elementary operation of a left corner parser is either a shift, an
attach or a left corner 
projection.  Under the independence assumptions mentioned above,
the probability of a shift will simply be the
probability of a certain left corner daughter ($lc$) being shifted given the
current goal category ($gc$), which we will model by $P_{shift}$.
Note that when to shift is deterministic.  If
a goal (i.e., minus) category is on top of the stack (and hence there is
no left corner category), then one must shift.  Otherwise one cannot.
If one is not shifting, one must choose to attach or project, which we
model by $P_{att}$.  Attaching only has a non-zero probability if the
left corner and the goal category are the same, but we define it for all
pairs.  If we do not attach, we project a constituent based on the left
corner with probability $P_{lc}$.
Thus the probability of
each elementary operation~$C_i$ can be expressed in terms of probability
distributions $P_{shift}$, $P_{att}$, and $P_{lc}$ as follows:
\begin{eqnarray*}
P(C_i = \shift lc) & = & \left\{ \begin{array}{ll}
P_{shift}(lc|gc) & \mbox{if top of the stack is $gc$} \\
0 & \mbox{otherwise}
\end{array} \right. \\
P(C_i = \attach) & = & \left\{ \begin{array}{ll}
P_{att}(lc, gc) & \mbox{if top of the stack is not $gc$} \\
0 & \mbox{otherwise}
\end{array} \right. \\
P(C_i = \project A \rightarrow \gamma) & = &
\left\{ \begin{array}{ll}
(1 - P_{att}(lc, gc))P_{lc}(A \rightarrow \gamma|lc, gc) & 
\mbox{if top of the stack is not $gc$} \\
0 & \mbox{otherwise}
\end{array} \right.
\end{eqnarray*}
Where these operations obey the following:
\begin{eqnarray*}
\sum_{lc} P_{shift}(lc|gc) & = & 1 \\
\mbox{If $lc \ne gc$, } P_{att}(lc, gc) & = & 0 \\
\sum_{\{A \rightarrow \gamma: \gamma = lc,\ldots\}} P(A \rightarrow
\gamma|lc, gc) & = & 1
\end{eqnarray*}
From the above we note that the probabilities of the choice of
projections sums to 
one, and hence, since other probabilities are complements of each other,
the probabilities of the actions available for each elementary operation
sum to one.  There are also no dead ends in a derivation, because unless
$A$ is a possible left corner constituent of $gc$, $P(A \rightarrow
\gamma|lc, gc) = 0$.  Thus we have shown that these probabilities define
a language model.%
\footnote{Subject to showing that the probability mass accumulates in
finite trees.}
%
That is, $\sum_s P(s|G) = 1$.
%
It is possible to
extend the PLCG model in various ways to include more probabilistic
conditioning, as we discuss briefly later, but our current results reflect
this model.

\section{Parsing experiments}

\subsection{PCFG Experiment}

Training and testing were done on Release 2 of the Penn Treebank
\cite{MaSaMa93}, published in 1995.  As in other recent work
\cite{Magerman95,Collins96}, training was done on sections 02--21 of the
Wall 
Street Journal portion of the treebank (approximately 40,000
sentences, 780,153 local trees) and final testing was done on section
23, which contains 2416 sentences.
Counts of how often each local tree occurred in the treebank were
made, and these were used directly to give probabilities for rewriting
each nonterminal.  The highest frequency rules are given in
\tabref{rulecounthigh}.
\begin{table}
\begin{center}
{\small
\begin{tabular}{llllc@{\hspace{2cm}}llllc}
\multicolumn{3}{l}{Rule} & Freq. & PCFG Prob. &
\multicolumn{3}{l}{Rule} & Freq. & PCFG Prob. \\[1ex]
PP & $\rightarrow$ & IN NP & 76617 & 0.81 &
NP & $\rightarrow$ & PRP & 17323 & 0.06 \\
NP & $\rightarrow$ & NP PP & 34965 & 0.11 & 
ADVP & $\rightarrow$ & RB & 14228 & 0.72 \\
NP & $\rightarrow$ & DT NN & 29351 & 0.09 &
NP & $\rightarrow$ & NN & 13586 & 0.04 \\
S  & $\rightarrow$ & NP VP & 28292 & 0.30 &
NP & $\rightarrow$ & NNS & 13318 & 0.04 \\
S  & $\rightarrow$ & VP & 23559 & 0.25 &
VP & $\rightarrow$ & TO VP & 12900 & 0.09 \\
S  & $\rightarrow$ & NP VP . & 17703 & 0.19 &
NP & $\rightarrow$ & NNP & 12575 & 0.04 \\
\end{tabular}}
\end{center}
\caption{Highest frequency CFG rules in Penn Treebank}\label{rulecounthigh}
\end{table}
Local trees were considered down to the level of
preterminals (i.e., part of speech tags); lexical information was
ignored.%
\footnote{Of course, we could easily integrate our model with
a tagging model.}
Every tree was given a new root symbol `ROOT', attached by a
unary branch to the root in the treebank.  Empty nodes (of which there
are several kinds in the treebank) were ignored,
and nonterminals above them that dominated no pronounced words were
also deleted.%
\footnote{Simply eliminating empties in the treebank is dangerous
because they are the only trace of unbounded dependency constructions.
This leads to ridiculous rules like \emph{S $\rightarrow$ VP} (with
23559 appearances in the treebank) stemming from \emph{S $\rightarrow$
NP VP} where 
there is a trace subject \emph{NP}.  A purely context-free solution would
be to introduce slash percolation.}
No attempt was made to do any smoothing.  While in a
few cases this would clearly be useful (e.g., the training data allows
a compound noun to be modified by four adjectives, but not a simple
noun), in practice the induced treebank grammar is hugely ambiguous,
and greatly overgenerates.  Thus, while the lack of smoothing in
principle disallows some correct parses from being generated, the
treebank grammar can always produce some parse for a sentence
\cite{Charniak96} and adding unseen rules with low probabilities is
unlikely to improve bottom line performance, because these added rules
are unlikely to appear in the maximum probability parse.  A better
solution would be to use a covering grammar with fewer rules and a
more deeply nested structure.

Testing was done by chart-parsing the part of speech tags of the
sentences (i.e., ambiguities in 
part of speech assignment were assumed to be successfully resolved).
An exhaustive chartparse was done and the highest
probability (Viterbi) parse was selected, in the standard way
\cite{Charniak93}.
Results from such parsing are shown in \tabref{pcfgres} together with
results from \cite{Charniak96}.
\begin{table}
\begin{center}
{\small
\begin{tabular}{@{}lrr@{}}
2--12 word sentences & this paper & Charniak\\[2ex]
Grammar (rules) & 14 971 & 10 605 \\
\% sent. length $<$ cutoff & 16.6\% \\
Test set size (sentences) & 401 \\
Average Length (words) &	8.3 & 8.7 \\
Precision &	89.8\% & 88.6 \\
Recall &	90.7\% & 91.7 \\
Labelled Precision &	83.5\%\\
Labelled Recall	& 82.9\%\\
Labelled Precision +1 & 87.1\% \\
Labelled Recall +1 & 85.2\% \\
Average CBs &	0.27\\
Non-crossing accuracy &	95.8\% & 97.9\%\\
Sentences with 0 CBs &	84.5\%
\end{tabular}}
\end{center}
\caption{PCFG results}
\label{pcfgres}
\end{table}
The measures shown have been used in various earlier works, and generally
draw from the PARSEVAL measures \cite{BlAbFlGdGrHaHiInJeKlLiMaRoSaSt91}.
Precision is how many 
brackets in the parse match those in the correct tree (perhaps also
examining labels), recall measures how many of the brackets in the correct
tree are in the parse.  The
unmarked measures ignore unary constituents, the ones marked +1
include unary constituents.\footnote
  {The\label{unaryfoot} original {\sc parseval} measures (because they were
   designed for 
   comparing different people's parsers that used different
   theories of grammar) ignored node labels entirely, discarded unary
   brackets, and performed other forms of tree normalization
   to give special treatment to certain cases such as verbal
   auxiliaries.  While such peculiarities made some sense in terms of
   the purpose for which the measures were originally developed, it is
   not clear they are appropriate for the use of these measures within
   the statistical NLP community.  Thus people often 
   report labelled measures, but it is often unclear whether
   the other rules and transformations of the standard
   have been employed
   (but this affects the results reported).  In this paper, unary nodes
   are deleted in measures except those marked +1, but none of the
   special case tree transformations in the {\sc parseval} standard
   are applied.  All punctuation and all constituent labels (but not
   functional tags) are also retained.} 
Crossing brackets (CBs) measures record how many
brackets in the parse cross bracketings in the correct tree, with the
non-crossing accuracy measuring the percent of brackets that are not CBs.
The `\% sent.\ length $<$ cutoff' says what percentage of sentences
within the 2416 sentence test section were shorter than the cutoff and
thus used in the test set for the current experiment.
Our results are not directly comparable to Charniak's
since he was using an earlier release of the Penn Treebank.  Further,
he used two strategies that aimed at increasing performance: recoding
auxiliaries from their Penn tag (which is undistinguished from other verbs) to
special auxiliary tags, and adding a (crude) correction factor to the
PCFG model so that it uniformly favored right-branching trees rather
than being context free.  Whether the former change was shown to be
beneficial is not discussed, but the later correction factor improved
results by about two percent.  The fact that the results are mixed
between the two systems suggests that the quality of the Penn Treebank
has improved in the second release, and these gains roughly match the
gains from these factors.  It is useful that the results are roughly
comparable since we can then use Charniak's results as a rough benchmark
for PCFG performance on longer sentences, which we have not
obtained.\footnote 
  {Our PCFG parser builds a complete chart, which leads to unviable
   space requirements for long sentences.}

Charniak's central contention is that purely structural parsing like
this using treebank grammars works much better than community lore would
have you believe.  Indeed, as the comparison in \tabref{charcoll} suggests, it
does not work much worse than \cite{Collins96}, a leading recent parser 
that includes lexical content.
\begin{table}
\begin{center}
{\small
\begin{tabular}{lrrr}
2/4--40 word sentences & PCFG & \footnotesize Magerman &
	 \footnotesize Collins\\[2ex] 
\% sent. length $<$ cutoff &	92.9\% \\
Test set size (sentences) & & 1759 & 2416 \\
Average Length (words) &	21.9 & 22.3 \\
Precision &	78.8\% & 86.3\%\\
Recall &	80.4\% & 85.8\%\\
Labelled Precision  & &	84.5\% & 86.3\%\\
Labelled Recall	& & 84.0\% & 85.8\%\\
Average CBs & &	1.33 & 1.14\\
Non-crossing accuracy &	87.7\%\\
Sentences with 0 CBs & & 55.4\% & 57.2\%
\end{tabular}}
\end{center}
\caption{PCFG \cite{Charniak96} vs.\
\cite{Magerman95}/\cite{Collins96} comparison} 
\label{charcoll}
\end{table}
That is, it seems one can score well in the PARSEVAL measures using
purely structural factors, and that the use of lexical factors in
other models is at present only adding a little to their
performance.\footnote
  {Again, results are not strictly comparable.  The comparison is
   unfair to Magerman and Collins' 
   systems since they are also doing part of speech tagging, whereas the
   PCFG is not.  But on the other hand, Magerman and Collins' parsers
   conflate the ADVP and PRT labels and ignore all punctuation, which
   improves their reported results.}
This is in part because the Penn treebank does not represent certain
semantic ``attachment'' decisions, and the structure of the trees
minimizes the penalty for other ``attachment'' errors, as we discuss in
the last section of this paper.

\subsection{LC parsing results}

The probabilistic left corner parser is implemented as a beam parser in C\@.
As a $k$-best beam parser, it is not guaranteed to find the best parse, unless
the beam is effectively infinite.
Space requirements depend on the size of the beam, and the length of the
sentence, but are considerably more reasonable than those for the chart
parser used above.  Nevertheless, the branching factor in the search
space is very high because there are many rules possible with a certain
left corner and goal category (especially for a grammar induced from the
Penn Treebank in the manner we have described).  Therefore, a huge beam
is needed for good 
results to be obtained.  A way of addressing this problem by binarizing
the grammar is discussed in the next section.  

The parser maintains two beams, one containing partial LC parses of the
first $i$ words, and another in which is built a beam of partial LC
parses of the first $i + 1$ words.   The partial parses are maintained
as pointers to positions in trie data structures that represent the list
of parser moves to this point and the current parse stack.  At the end
of parsing, the lists of parser moves can be easily turned into parse trees
for the $n$-best parses in the beam.  

Results are shown in \tabref{lcresults}.  They reflect the
same training and testing data as described above.
The results show a small increase in performance for PLCGs.  This is shown
more dramatically in the right-hand column of \tabref{lcresults}, which shows
that the extra 
information provided by the Left Corner goal category reduces parsing errors
by about 20\% over our PCFG results on 2--12 word sentences.
\begin{table}
\begin{center}
{\small
\begin{tabular}{@{}lrrrr@{}}
& \multicolumn{3}{c}{Parser results} & 2--12 Error \\
Sentence Lengths: & 2--12 & 2--16 & 2--25 & Reduction \\
Beam size & 50 000 & 50 000 & 40 000\\
\% sent. length $<$ cutoff &	16.6\% & 28.1\% & 60.2\%\\
Test set (sentences) & 		401    & 680    & 1454\\
Average length (words) &	8.3    & 10.9   & 16.3\\[1ex]
Precision &			92.0\% & 90.1\% & 84.6\% & 21.6\% \\
Recall &			92.3\% & 89.5\% & 83.2\% & 17.2\% \\
Labelled Precision &		87.1\% & 86.0\% & 81.1\% & 21.8\% \\
Labelled Recall	& 		86.7\% & 84.9\% & 79.6\% & 22.2\% \\
Labelled Precision +1 & 	88.6\% & 87.7\% & 83.5\% & 11.6\% \\
Labelled Recall +1 & 		88.3\% & 86.3\% & 81.5\% & 20.9\% \\
Average CBs &			0.21   & 0.43   & 1.25   & 22.2\% \\
Non-crossing accuracy &		96.8\% & 94.7\% & 89.6\% & 23.8\% \\
Sentences with 0 CBs &		87.5\% & 76.0\% & 52.0\% & 19.4\%
\end{tabular}}
\end{center}
\caption{Left Corner Parser results from $n$-ary grammar.}
\label{lcresults}
\end{table}


\section{Binarization}

For the parser above, use of a beam search is quite inefficient because, for a
given left corner and goal category, there are often hundreds of possible
local trees that could be projected, and little information is available
at the time this decision is made, since the decision mainly depends on
words that have not yet been shifted.  Therefore the
beam must be large for good results to be obtained, and, at any rate, the
branching factor of the search space is extremely high, which slows
parsing.  One could imagine using various heuristics to improve the
search, but the way we have investigated combatting this problem is by
binarizing the grammar. 

The necessary step for binarization is to eliminate productions with three
or more daughters.  We carried this out by merging the tails, so that
a rule such as \emph{NP~$\rightarrow$~Det~JJ~NN} is replaced by two rules,
\emph{NP~$\rightarrow$~Det~NP$_{Det}$} and
\emph{NP$_{Det}$~$\rightarrow$~JJ~NN}.  This is  
carried out recursively until only binary rules remain.  As a result of
this choice of binarization, $n$-ary rules that share
the same mother and left corner category all reduce to a single rule.
This greatly cuts the branching factor of the search space and 
allows decisions to be put off during
parsing, until more of the input has been seen, at which point
alternative continuations can be better evaluated.  Furthermore, the
weights for such rules all combine into a larger weight 
for the combined rule.%
\footnote{If we were to do this even for rules that start out binary,
and eliminate unary rules downward rather than upward, then for every
left corner $C$ and mother $G$ there will be a unique rule 
$G \rightarrow C\ G_C$.}

It is important to note that the resulting model is \emph{not}
equivalent to our original model.  While the straightforward way of
binarizing a PCFG yields the same probability estimates for trees as the
$n$-ary grammar, this is not true for our PLCG model since we are now
introducing new estimates for shifting terminals for each of our
newly created non-terminals.  Slightly different probability estimates
result, and further work is needed to investigate what relationship
exists between them and the probability estimates of the original grammar.

Prior to the above binarization step, one might also wish to eliminate
unary productions, much as we earlier eliminated
empty categories.  This can be done in two ways.  One way is to fold
them upwards.  This 
preserves lexical tagging.  That is, if there is a category $A$
dominating only a tree rooted at $B$, then the category $A$ is
eliminated and the tree rooted at $B$ moved upwards.  This may cause
the number of rules to increase, because a local tree that started
with a daughter $A$ will now show up with daughter $B$ in the same
place.  The alternative is to eliminate $B$ and replace
it with $A$.  This can also create a new local tree instance because
the daughters of $B$ now show up with a new mother $A$.  In this way,
lexical tags can be changed.  For instance, consider a rule $NP
\rightarrow NNP$ for a noun phrase rewriting as a proper noun.  In the
context of eliminating empty categories upward, we get a new rule $S
\rightarrow NNP\ VP$, whereas by eliminating empty categories downward
we would have produced a new lexical entry $NP \rightarrow Jones$.  Our
current results do not reflect the elimination of unary rules, but we
hypothesize that doing this would further improve the measures that do
not consider unary nodes, while probably harming the results on measures
that do include unary nodes.

\begin{table}
\begin{center}
{\small
\begin{tabular}{@{}lrrrrr@{}}
 & \multicolumn{4}{c}{Parser Results} & 2--25 Error \\
Sentence Lengths:      &  2--12 & 2--16  & 2--25  & 2-40   & Reduction\\
Beam size              & 40 000 & 40 000 & 40 000 & 40 000 \\
\% sent. length $<$ cutoff & 
			 16.3\% & 28.1\% & 60.2\% & 91.8\% \\
Test set (sentences)   &        &  680   & 1454  &  2216 \\
Average length (words) &    8.3 & 10.9   & 16.3   &   21.6 \\[1ex]
Precision              & 93.5\% & 91.4\% & 86.6\% & 83.0\% & 13.0\% \\
Recall                 & 92.8\% & 89.9\% & 84.3\% & 80.7\% &  6.5\% \\
Labelled Precision     & 89.8\% & 88.1\% & 83.4\% & 79.9\% & 12.2\% \\
Labelled Recall        & 88.4\% & 86.3\% & 81.0\% & 77.6\% &  6.9\% \\
Labelled Precision +1  & 90.0\% & 89.0\% & 85.1\% & 81.9\% &  9.7\% \\
Labelled Recall +1     & 89.5\% & 87.2\% & 82.6\% & 79.5\% &  5.9\% \\
Average CBs            & 0.17   & 0.39   & 1.09   & 1.99   & 12.8\% \\
Non-crossing accuracy  & 97.2\% & 95.2\% & 90.9\% & 87.6\% & 12.5\% \\
Sentences with 0 CBs   & 89.8\% & 78.2\% & 55.8\% & 41.5\% &  7.9\%   
\end{tabular}}
\end{center}
\caption{Left Corner Parser results with binarized grammar}
\label{binresults}
\end{table}
\tabref{binresults} shows that binarization brings a further modest
improvement in results.  The righthand column shows the percent error
reduction on 2--25 word sentences between the $n$-ary grammar and the
binary grammar.

\section{Extended Left Corner Models}

More sophisticated PLCG parsing models will naturally provide greater
conditioning of the probability of an elementary operation based on the
parse history.  There are a number of ways that one could then proceed.
From the background of work on LC parsing, a natural factor to consider
is the size of the parse stack, and we will briefly investigate
incorporating this factor.

For left corner parsing, the stack size is particularly interesting.
\cite{Stabler94} notes that in contrast to bottom-up and top-down
parsing methods, left-corner parsers can handle arbitrary left-branching
and right-branching structures with a finitely bounded stack 
size.  Furthermore, left-corner parses of center embedded
constructions have stack lengths proportional to the amount of
embedding.  This is not actually true for the stack-based LC parser
presented earlier which has the stack growing 
without bound for rightward-branching trees.  To gain the desirable
property that Stabler notes, one needs to do stack composition by
deciding immediately whether to attach whenever one projects a category
that matches the current goal category, rather than delaying the
attachment until after the left corner constituent is complete.
If one decides to attach,
the goal minus category and the mother category are
immediately removed from the stack.  This is implemented by replacing
the predicate \verb+process+ in \figref{lcstackprog} by the code in
\figref{lcstacksmall}, where composition is done by the predicate
\verb+compose+.  
\begin{figure}
{\small
\begin{verbatim}
process([Item|Items], Stack, L, L) :- % lc
    rule(LHS, [Item|Rest]),
    compose(LHS, Items, Stack1),
    predict(Rest, Stack1, Stack).
process(Stack, [L|Stack], [L|Ls], Ls). % shift

compose(A, [m(A)|L], L). % attach
compose(A, L, [A|L]).    % don't
\end{verbatim}}
\caption{Altered code for a Prolog LC parser that does stack
composition}\label{lcstacksmall} 
\end{figure}
All else being equal, this change makes no difference to the
probabilistic model presented earlier.  In practice though, this
formulation makes a beam search less effective since we are bringing
forward the decision of whether to attach or not, and often both
alternatives must be tried which fills out the beam unnecessarily.

Given human intolerance for center embeddings and ease
of parsing left and right branching structures, we would expect the
stack sizes to stay low.  The general prediction (and 
empirical fact) is that the probability of not attaching (composing) decreases
slightly with the size of the stack.   The counts for the Penn Treebank,
after binarization (including removal of unary rules), are given in 
\tabref{binstack}.  Once one looks beyond the odd-even effect in the
data, the decreasing probability of not attaching can be clearly seen.
\begin{table}
\begin{center}
\begin{tabular}{rrrrr}
Stack & &\multicolumn{3}{c}{Stack size change} \\
Size  & Total & $-1$ & 0 & +1 \\[1ex]
8 & 23 & 20 (87\%)& 3 (13\%) & 0 \\
7 & 160 & 86 (54\%) & 54 (34\%) & 20 (13\%) \\
6 & 1291 & 987 (76\%) & 218 (17\%) & 86 ( 7\%) \\
5 & 3745 & 1393 (37\%) & 1365 (36\%) & 987 (26\%) \\
4 & 17000 & 12116 (71\%) & 3491 (21\%) & 1393 (8\%) \\
3 & 39105 & 12241 (31\%) & 14748 (38\%) & 12116 (31\%) \\
2 & 108544 & 63160 (58\%) & 33143 (31\%) & 12241 (11\%) \\
1 & 71681 & 8521 (12\%) & 0 $\phantom{(11\%)}$ & 63160 (88\%)
\end{tabular}
\end{center}
\caption{Changes in stack size for Penn Treebank}\label{binstack}
\end{table}

To incorporate stack length into the model, we wish to more accurately
predict: $P(C_i|C_1, \ldots, C_{i-1})$.
Previously, histories of parse steps were equivalenced
according to just the goal category and the left corner (if present).
Now, we are going to additionally differentiate parse histories
depending on $\ell$, the length of the stack after 
$C_{i-1}$.  As before, if the top of
the stack is a predicted category, we will shift; otherwise we cannot
shift.  In the latter case, we will predict $P_{\delta}(\delta| \ell,
gc, lc)$, the probability of various changes in 
the stack size, based on the stack size, the goal category, and the left
corner.  

Let us assume that rules
are binary, as discussed above.  Then the
possible change in the stack length from a single elementary operation is
between $-2$ and +1.  Given that we 
are not shifting, we have the following:
\begin{quote}
\begin{tabular}[t]{rl}
Stack Delta & Rule Type \\
$-2$ & unary left corner and attach\\
$-1$ & binary left corner and attach\\
0 & unary left corner (no attach)\\
+1 & binary left corner (no attach)
\end{tabular}
\end{quote}
The probability of each elementary operation will then be the
probability of a certain stack delta (given the stack size, left corner
and goal category) times the
probability of a certain rule, given the left corner, the goal category,
and the stack delta.  Whether we attach (compose) or not is
deterministic given the stack delta, and so we no longer need to model
the $P_{att}$ distribution.  Under this model, the probability of
different projection operations (given that we are not in a position to
shift) becomes: 
\begin{eqnarray*}
P(C_i = \project A \rightarrow lc, \mbox{ attach}) & = &
P_{\delta}(-2|\ell, lc, gc)P_{lc'}(A \rightarrow lc|lc, gc,
 \ell, \delta) \\
P(C_i = \project A \rightarrow lc, \mbox{ do not attach}) & = &
P_{\delta}(0|\ell, lc, gc)P_{lc'}(A \rightarrow lc|lc, gc,
 \ell, \delta) \\
P(C_i = \project A \rightarrow lc\ c_2, \mbox{ attach}) & = &
P_{\delta}(-1|\ell, lc, gc)P_{lc'}(A \rightarrow lc\ c_2|lc, gc,
 \ell, \delta) \\
P(C_i = \project A \rightarrow lc\ c_2, \mbox{ do not attach}) & = &
P_{\delta}(+1|\ell, lc, gc)P_{lc'}(A \rightarrow lc\ c_2|lc, gc,
 \ell, \delta)
\end{eqnarray*}

A variety of other extended models are possible.  Note that the model
does not need to be uniform, and that we can estimate different
classes of elementary operations using different probabilistic
submodels.  In particular, at the level of preterminals, we can
incorporate a tagging model.  Given the structure of the Penn
Treebank, a terminal $w_j$ is always dominated by a unary rule giving
the terminal's part of speech $p_j$. In line with our basic model
above, the choice of part of speech for a word (where the word now
counts as the left corner) will certainly depend on the current goal
category.  However, we can also condition it on other preceding parse
decisions, in particular on the part of speech of the preceding two
words, or, perhaps, in certain circumstances, on the particular word
that preceded.  Taking the former possibility, we can say, for cases
where $C_i$ involves predicting a preterminal (through a left corner
projection step),
\begin{eqnarray*}
P(C_i|C_1, \ldots, C_{i-1}) & = & P(p_j|w_j, gc, C_1, \ldots, C_{i-1}) \\
 & \approx & P(p_j|w_j, gc, p_{j-2}, p_{j-1})
\end{eqnarray*}
Assuming -- perhaps rashly -- independence between the conditioning
variables that we have been using before ($lc$, $gc$) and the new ones
($p_{j-1}$, $p_{j-2}$), then we have that:
\[
P(p_j|w_j, gc, p_{j-2}, p_{j-1}) \approx
\frac{P(p_j|w_j, gc)P(p_j|p_{j-2}, p_{j-1})}{P(p_j)}
\]
This is nice in part because we can calculate it using just the statistics
previously gathered and a simple trigram model over POS tags.
(An incidental nice property is that the probability is for 
the POS given the word, and not the other way round, as in the
`confusing' $P(w|t)$ term of standard Markov model POS taggers.)%
\footnote{Below is a derivation of this equation, written in a
simplified form with 
three variables.  We assume $b$ and $c$ are independent, and the result
then follows by using Bayes' rule followed by the definition of conditional
probability:
\[
P(a|b,c) = \frac{P(a)P(b,c|a)}{P(b,c)} = 
 \frac{P(a)P(b|a)P(c|a,b)}{P(b)P(c|b)}
 = \frac{P(a)P(b|a)P(c|a)}{P(b)P(c)}
 = \frac{P(a)P(b,a)P(c,a)}{P(a)^2P(b)P(c)} 
= \frac{P(a|b)P(a|c)}{P(a)}
\]
}

\section{Comparison with previous work}

Previous work on non-lexicalized parsing of Penn Treebank data includes
\cite{ScRoOs93} and \cite{Charniak96}, but the work perhaps most
relevant to our own is that of \cite{BrCa93} and \cite{CaBr96}, which
also seeks to address the context-freeness assumption of
PCFGs.  
They approach the problem by using a probabilistic model based on LR
parse tables.  Unfortunately, many differences of approach make
meaningful comparisons difficult, and a comparable study of using PLCGs
versus Probabilistic LR parsing remains to be done.  Briscoe and Carroll
use their Probabilistic LR grammar to guide the actions of a
unification-based parser which uses a hand-built grammar.  While we are
sympathetic with their desire to 
use a more knowledge-based approach, this means that: their language
model is deficient, since probability mass is given to derivations which
are ruled out because of unification failures; the coverage of their
parser is quite limited because of limitations of the grammar used; and
much time needs to be expended in developing the grammar, whereas our
grammar is acquired automatically (and quickly) from the treebank.
Moreover, while results are not directly comparable, our parsers seem
to do rather better on precision and recall than the parser described in
\cite{CaBr96}, while performing somewhat worse on crossing brackets
measures.  However, Carroll and Briscoe's inferior results probably
reflect the fact that the parse trees of their grammar do not match
those of their treebank more than anything else.

\section{Observations on why parsing the Penn Treebank is easy}

How is it that the purely structural -- and context free even in
structural terms -- PCFG parser manages to perform so well?
An important observation is that the measures of precision and recall
(labelled or not) and crossing brackets are actually quite easy measures to
do well on.  It is important to notice that they are measuring success at
the level of individual decisions -- and normally what makes NLP hard is
that you have to make many consecutive decisions correctly to succeed.  The
overall success rate is then the $n^{th}$ power of the individual decision
success rate -- a number that easily becomes small.

But beyond this, there are a number of features particular to the structure
of the Penn Treebank that make these measures particularly easy.  Success
on crossing brackets is helped by the fact that Penn Treebank trees are
quite flat.  To the extent that sentences have very few brackets in them,
the number of crossing brackets is likely to be small.  Identifying
troublesome brackets that would lower precision and recall measures is also
avoided.  As a concrete instance of this, one difficulty in parsing is
deciding the structure of noun compounds \cite{Lauer95}.  Noun compounds of
three or more words in length can display any combination of left- or
right-branching structure, as in [[cable modem] manufacturer] vs.
[computer [power supply]].  But such fine points are finessed by the Penn
Treebank, which gives a completely flat structure to a noun compound (and
any other pre-head modifiers) as shown below (note that the first example also
illustrates the rather questionable Penn Treebank practice of tagging
hyphenated non-final portions of noun compounds as adjectives!).

\begin{center}
\begin{tabular}{l}
(NP a/{\sc dt} stock-index/{\sc jj} arbitrage/{\sc nn} sell/{\sc nn}
 program/{\sc nn} )\\
(NP a/{\sc dt} joint/{\sc jj} venture/{\sc nn} advertising/{\sc nn}
 agency/{\sc nn} )
\end{tabular}
\end{center}

	Another case where peculiarities of the Penn Treebank help is the
(somewhat nonstandard) adjunction structures given to post noun-head
modifiers, of the general form (NP (NP the man) (PP in (NP the moon))).
A well-known parsing ambiguity is whether PPs attach
to a preceding NP or VP -- or even to a higher preceding node -- and
this is one for which lexical or contextual information is clearly
much more important than structural factors \cite{HiRo93}.
Note now that the use of the above adjunction structure reduces the 
penalty for making this decision wrongly.  Compare Penn Treebank style
structures, and another common structure in the examples in
\tabref{pennvother}.
\begin{table}
\begin{center}
\begin{tabular}{@{}ll@{}}	
Penn VP attach &	(VP saw (NP the man) (PP with (NP a telescope))) \\
Penn NP attach &	(VP saw (NP (NP the man) (PP with (NP a telescope))))\\
Another VP attach &	(VP saw (NP the (N$'$ man)) (PP with (NP a (N$'$
telescope))))) \\
Another NP attach &	(VP saw (NP the (N$'$ man (PP with (NP a (N$'$
telescope))))))
\end{tabular}
\end{center}
\caption{Penn trees versus other trees}
\label{pennvother}
\end{table}
Note the difference in the results:
\begin{center}
\begin{tabular}{@{}llccc@{}}

 & Error & \multicolumn{3}{c}{Errors assessed} \\
 &	& Prec. &	Rec. &	CBs\\[1ex]
Penn & VP instead of NP &	0 &	1 &	0 \\
 & NP instead of VP &	1 &	0 & 	0 \\
Another & VP instead of NP & 	1 & 	2 &	1 \\
& NP instead of VP & 	2 & 	1 &	1
\end{tabular}
\end{center}
The forgivingness of the Penn Treebank scheme is manifest.  One can
get the attachment wrong and not have any crossing brackets.\footnote
  {If one includes unary brackets (recall footnote~\ref{unaryfoot}),
   then the contrast becomes even more marked, since there would be 2
   precision and recall errors each under the alternative parse
   trees.}

\section{Conclusions}

This paper explores a new class of probabilistic parsing algorithms for
context-free grammars, probabilistic left corner grammars.  The ability
of left corner parsers to support left-to-right online parsing makes
them initially promising for many tasks.  The
different conditioning model is slightly richer than that of standard
PCFGs, and this was shown to bring worthwhile
performance improvements over a standard PCFG when used to parse Penn
Treebank sentences.  Beyond this, the model can be extended
in various ways, an avenue that we have only
just begun exploring.  Because the left-corner component of the
grammar is purely structural, it can be combined with other models
that include lexical attachment preferences and preferences for basic
phrasal chunking (both incorporated into Collins' parser).

\bibliographystyle{apalike}


\end{document}